\documentclass[sigconf,nonacm]{acmart}

\usepackage{xspace}
\usepackage{tikz}
\usepackage{adjustbox}
\usepackage{pifont}

\newcommand{\sys}{HyperPose\xspace}
\newcommand{\eg}{\emph{e.g.,}\xspace}

\newcommand*\myc[1]{%
\scalebox{0.78}{\begin{tikzpicture}[baseline=-3pt]
  \node[draw,circle,inner sep=0.5pt, fill=black] {\textcolor{white}{\textsf{\textbf{#1}}}};
\end{tikzpicture}}}
\newcommand{\cmark}{\ding{51}}%
\newcommand{\xmark}{\ding{55}}%
\newcommand{\tabincell}[2]{\begin{tabular}{@{}#1@{}}#2\end{tabular}}

\AtBeginDocument{%
  \providecommand\BibTeX{{%
    \normalfont B\kern-0.5em{\scshape i\kern-0.25em b}\kern-0.8em\TeX}}}

\begin{document}
\fancyhead{}

\title{Fast and Flexible Human Pose Estimation with HyperPose}

\author{Yixiao Guo}
\authornote{Equal contributions. Yixiao implemented the model development library. Jiawei implemented the model inference engine. Guo prototyped the initial version of HyperPose.}
\affiliation{%
  \institution{{Peking University}}
   \state{Beijing}
  \country{{China}}
  }

\author{Jiawei Liu}
\authornotemark[1]
\affiliation{%
  \institution{{Tongji University}}
   \state{Shanghai}
  \country{{China}}
  }

\author{Guo Li}
\authornotemark[1]
\affiliation{%
  \institution{{Imperial College London}}
   \city{London}
  \country{{United Kingdom}}
  }

\author{Luo Mai}
\affiliation{%
  \institution{{University of Edinburgh}}
   \city{Edinburgh}
  \country{{United Kingdom}}
  }

\author{Hao Dong}
\affiliation{%
  \institution{{Peking University}}
  \state{Beijing}
  \country{{China}}
  }

\renewcommand{\shortauthors}{Y. Guo, J. Liu, G. Li, L. Mai, H. Dong}

\begin{abstract}
Estimating human pose is an important yet challenging task in multimedia applications. 
Existing pose estimation libraries target reproducing standard pose estimation algorithms.
When it comes to customising these algorithms for real-world applications, none of the existing libraries can offer both the flexibility of developing custom pose estimation algorithms and the high-performance of executing these algorithms on commodity devices. 
In this paper, we introduce \sys, a novel flexible and high-performance pose estimation library. \sys provides expressive Python APIs that enable developers to easily customise pose estimation algorithms for their applications.
It further provides a model inference engine highly optimised for real-time pose estimation. This engine can dynamically dispatch carefully designed pose estimation tasks to CPUs and GPUs, thus automatically achieving high utilisation of hardware resources irrespective of deployment environments. 
Extensive evaluation results show that \sys can achieve up to 3.1x$\sim$7.3x higher pose estimation throughput compared to state-of-the-art pose estimation libraries without compromising estimation accuracy. 
By 2021, \sys has received over 1000 stars on \href{https://github.com/tensorlayer/hyperpose}{GitHub} and attracted users from both industry and academy.\footnote{The published version of this work is at https://doi.org/10.1145/3474085.3478325}

\end{abstract}


\maketitle

\section{Introduction}

Multimedia applications, such as interactive gaming, augmented reality and self-driving cars, can greatly benefit from accurate and fast human pose estimation.
State-Of-The-Art (SOTA) pose estimation algorithms (\eg OpenPose~\citep{openpose}, PoseProposal~\citep{poseproposal}, and PifPaf~\citep{kreiss2019pifpaf})
pre-process video streams, use neural networks to infer human anatomical key points, and then estimate the human pose topology.

In practice, achieving both high accuracy and high-performance in pose estimation is however difficult. 
On the one hand, achieving high-accuracy pose estimation requires users to deeply customise standard pose estimation algorithms (e.g., OpenPose and PifPaf), so that these algorithms can accurately reflect the characteristics of user-specific deployment environments (e.g., object size, illumination, number of humans), thus achieving high accuracy. 
On the other hand, deeply customised pose estimation algorithms can contain various \emph{non-standard} computational operators in pre-processing and post-processing data, making such algorithms difficult to always exhibit high performance in using commodity embedded platforms (e.g., NVIDIA Jetson and Google TPU Edge)

When using existing pose estimation libraries to develop custom applications, users often report several challenges. High-performance C++-based libraries such as OpenPose~\citep{openpose} and AlphaPose~\citep{alphapose} focus on specific pose estimation algorithms.
They do not provide intuitive APIs for users to customise pose estimation algorithms based on the requirements of their specific deployment environments.
These libraries are also optimised for certain hardware platforms. When re-targeting them to new hardware platforms, users must largely modify their internal execution runtime, which is non-trivial for most pose estimation algorithm developers. 
Furthermore, users could also use high-level pose estimation libraries such as TF-Pose~\citep{tf-pose} and PyTorch-Pose~\citep{pytorch-pose}.
These libraries offer users with Python APIs to easily declare various pose estimation algorithms. However, this easiness comes with a large performance overhead, making these libraries incapable of handling real-world deployment where high-resolution images are ingested at high speed~\cite{xu2021move}. 

In this paper, we introduce \sys, 
a flexible and fast library for human pose estimation.
The design and implementation of \sys makes the following contributions:

\noindent
\textbf{(i)~Flexible APIs for developing custom pose estimation algorithms.} 
\sys provides flexible Python APIs for developing custom pose estimation algorithms. These APIs consist of those for customising the pipelines of pose estimation algorithms, the architectures of  deep neural networks, training datasets, training hyper-parameters~\cite{mai2019taming}, data pre-processing pipelines, data post-processing pipelines, and the strategy of paralleling neural network training on GPUs. We show that,
using these APIs, users can declare a wide range of commonly used pose estimation algorithms while customising these algorithms for high estimation accuracy. 

\noindent
\textbf{(ii)~High-performance model inference engine for executing custom pose estimation algorithms.}
\sys can achieve high-performance in executing custom pose estimation algorithms. This is achieved through a novel \emph{high-performance algorithm execution engine}. This engine 
is designed as a streaming dataflow~\cite{mai2018chi}.  
This dataflow can take custom computational operators for implementing custom pose estimation logic. These operators can be dynamically dispatched onto parallel CPUs and GPUs, thus keeping computational resources always busy, irrespective of model architectures and hardware platforms. The implementations of these operators are also highly optimised, mainly by carefully leveraging the optimised computer-vision library: OpenCV, and the high-performance model inference library: TensorRT.

We study both the API easiness and the performance of \sys.
The API study shows that
\sys can provide 
better flexibility via building and customising pose estimation algorithms.
The test-bed experiments further show that \sys can out-perform
the state-of-the-art optimised pose estimation framework: OpenPose by to up to 3.1x in terms of the processing throughput of high-resolution images.

\section{Design and Implementation}
In this section, we present the design principles and implementation details of \sys.

\subsection{Expressive Programming APIs}
\sys aims to support different types of users in developing pose estimation algorithms. 
There are users who would like to find  suitable algorithms and adapt them for their applications. To support this, \sys allows users to customise the pipeline of a typical pose estimation algorithm. 
Other users would like to further modify the components in a pose estimation pipeline. For example, they often need to control how a deep neural network is being trained, and the data pre-processing/post-processing operators being used. To support them, \sys allow users to plug in user-defined components in a pose estimation pipeline.

\subsubsection{Algorithm development APIs}
\sys provides a set of high-level APIs~\footnote{\url{https://hyperpose.readthedocs.io/en/latest/}} to relieve users from the burden of assembling the complex pose estimation system. The APIs are in three modules, including \emph{Config}, \emph{Model}, and \emph{Dataset}. \emph{Config} exposures APIs to configure the pose estimation system, while \emph{Model} and \emph{Dataset} offer APIs to construct the concrete model, dataset, and the development pipeline. With each API, users can configure the architecture for different algorithms (\eg  OpenPose, PoseProposal), backbone networks (\eg  MobileNet \citep{mobilenetv1} and ResNet\citep{resnet}), 
and the training or evaluation datasets (\eg  COCO and MPII). 
For the development procedure,
users can configure the hyper-parameters (\eg  learning rate and batch size), the distributed training via the KungFu library~\cite{mai2020kungfu} option (\eg using
a single or multiple GPUs), the training strategy (\eg adopting pre-training and adaptation stage), and 
the format of storing a trained model for further deployment(\eg ONNX and TensorRT UFF).
The rich configuration options 
enable users to efficiently adapt the off-the-shelf models.

\subsubsection{Algorithm customisation APIs}
\sys users can flexibly customise pose estimation algorithms to best fit in with their specific usage scenarios.
This is achieved by providing common interfaces for key
components in the algorithms.
For example, to implement custom neural networks,
users could inherit from the \texttt{Model} class defined in \sys,
and warping the custom computation logic into the corresponding member functions.
As long as use the self-defined model to replace the preset model options during configuration, the custom model is enabled. The same practice
applies to enabling a custom dataset. 
These customised components are then automatically
integrated into the pose estimation system by \sys. 
The \emph{Model} module further exposures processing modules including \texttt{preprocessor}, \texttt{postprocessor} and \texttt{visualizer},
which allow users to assemble their own development pipeline.
By doing this, \sys makes its APIs flexible to support
extensive customisation of its pose estimation algorithms.

\subsection{High-performance Execution Engine}

\begin{figure*}[h]
    \centering
    \includegraphics[width=\linewidth]{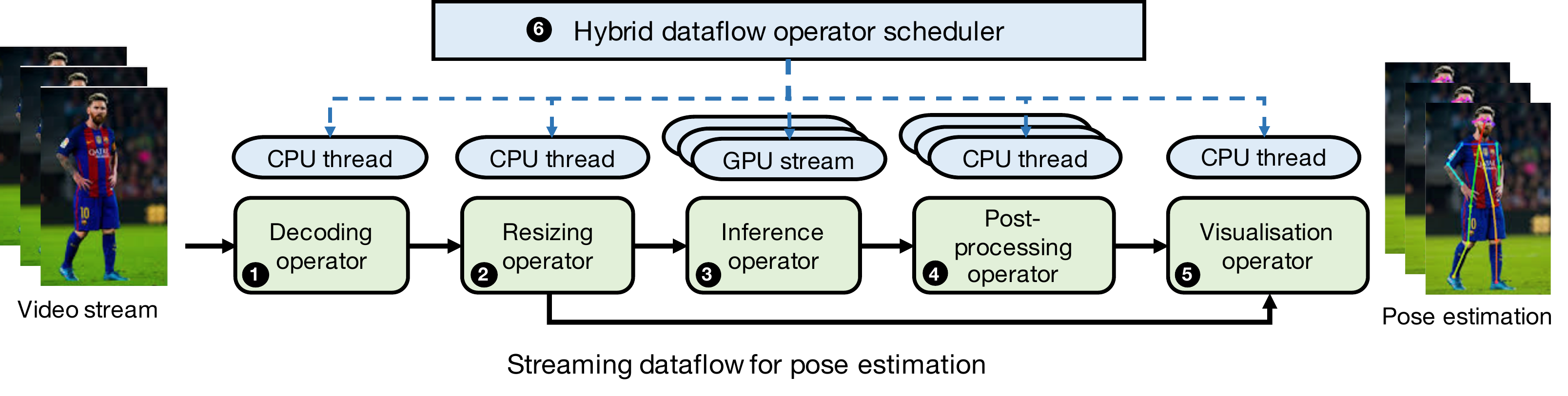}
    \caption{Architecture of the \sys C++ execution engine.}
    \label{fig:architecture}
    \Description{\sys C++ execution engine models pose estimation algorithms as a data stream pipeline, abstracts the common computation of pose estimation algorithms as dataflow operators and uses a hybrid dataflow operator scheduler for maximising performance.}
\end{figure*}

Designing a high-performance execution engine for human pose estimation is challenging.
A human pose estimation pipeline consists of video stream ingesting, pre-processing (\eg resizing and data layout switching), GPU model inference, CPU post-processing, and result exportation (\eg visualisation).

These computation steps must be collaboratively completed using heterogeneous devices:
CPUs, GPUs, and I/O devices (\eg disk, cameras, 
etc.). 
To achieve real-time inference, 
the engine must maximise the efficiency of using \emph{all} these devices through parallelism. 
We make two key designs in our inference engine:

\subsubsection{Streaming pose estimation dataflow}
We abstract the computation
steps shared by most pose estimation algorithms 
and implement these steps as
\emph{dataflow operators}.
The operators can be asynchronously executed in a streaming dataflow. 
The topology of the dataflow is implemented to be static for provisioning better optimisation, thus maximising the processing throughput.

Figure~\ref{fig:architecture} illustrates the dataflow implemented in \sys.
The source for the dataflow is often a video stream produced by a real-time device (\eg camera). 
The dataflow ingests the images into
a \emph{decoding operator} (see \myc{1}).
The decoded images are resized and transformed to an expected data layout (\eg channel-first) so that they can be fed into a neural network (see \myc{2}).
This neural network is executed by an \emph{inference operator} (see \myc{3}) on GPU.
This operator loads the checkpoint of a
neural network trained by the \sys Python platform or other supported libraries. 
The computed activation map from the neural network is given to a \emph{post-processing operator} for
parsing human pose topology (see \myc{4}). 
The topology is placed onto the original image
at a \emph{visualisation operator} (see \myc{5}). 

Each operator pair shares a concurrent FIFO channel, manipulated by CPU threads.
The operators only block when there is no incoming record and sleep until they are notified, thanks to the condition variable mechanism.
Benefiting from this, \sys can ensure the pose estimation dataflow fully utilise parallel heterogeneous processors including CPUs and GPUs. 
Besides,
\sys fully masks I/O latency (\eg waiting for images
from a camera) by overlapping I/O operations with computation operations.

\subsubsection{Hybrid dataflow operator scheduler}
We design a \emph{hybrid dataflow operator scheduler}
in the execution engine (see \myc{6} in Figure~\ref{fig:architecture}), for further improving the CPU/GPU utilisation in addition to the pipeline parallelism.
Regarding GPU utilisation, we leverage dynamic batching in front of the inference operator, encouraging the inference operator to take a larger batch as input each time.
In concert with the streaming dataflow mechanism, the batching slot only accumulates more input tensors when the GPU becomes the bottleneck.
Such optimisation is beneficial since i) batching reduces the times of GPU kernel launch thus improving GPU processing throughput, ii) when the GPU is the bottleneck, batching gets enhanced to alleviate the congestion~\cite{koliousis12crossbow}, and iii) when the GPU is not the bottleneck, batching gets weakened for not deteriorating the per-image response delay.
As to CPU threads scheduling, we implemented an asynchronous thread-level communication mechanism based on conditional variables.
When the bottleneck happens in a CPU-based operator (\eg the post-processing operator), the working threads of non-blocking operators will fall asleep to save CPU cycles for the bottleneck until the next round starts.

\subsection{Implementation and Compatibility}

\sys supports 3 classes of pose estimation algorithms: 1) PAF\citep{openpose} (i.e., OpenPose), 2) PoseProposal\citep{poseproposal}, and 3) PifPaf\citep{kreiss2019pifpaf}.

Since developing and deploying pose estimation algorithms have different objectives, \sys separates the implementation for training and inference but makes their ecosystem well-compatible.

The training library is a Python library implemented using Tensorflow and TensorLayer~\cite{tensorlayer} for DNN construction with Numpy and other common libraries for post-processing.

For maximum performance, the whole inference engine is notably implemented in C++ 17 for massive low-level code optimisation and parallelism.
The GPU inference operator is based on NVidia TensorRT, one of the fastest DNN inference libraries.
The imaging-related operations are based on OpenCV,
and the dataflow scheduler is implemented by the C++ standard thread library.

The implementation of \sys is compatible with many pose estimation algorithms, such as DEKR~\cite{Geng2021CVPR} and CenterNet~\cite{zhou2019objects}. These algorithms share the bottom-up architectures as those (e.g., OpenPose and PifPaf) implemented in 
\sys. They are thus easy to be implemented as extensions to \sys. 

\section{Evaluation}
\begin{table}
    \caption{API study. * denotes single-human datasets only.}
    \label{tab:api_comparison}
    \begin{tabular}[width=\columnwidth]{llllll}
        \toprule
                     &Algorithm  & Dataset   & DNN   & Config. & Ext.   \\
        \midrule
        TF-Pose      &  1        &  1        &  5    & 5       & \xmark \\ 
        PyTorch-Pose &  2        &  3*       & 13    & 26      & \xmark \\ 
        \textbf{HyperPose} & 3   & 2         & 10    & 30      & \cmark \\ 
        \bottomrule
    \end{tabular}
\end{table}

\begin{table}
    \caption{Performance Evaluation of  Inference Engine.   \protect\footnotemark}
    \label{table:perf_bench}
    \begin{tabular}[width=\textwidth]{cccc}
        \toprule
         Configuration &
         \tabincell{c}{Baseline\\FPS} & \tabincell{c}{Our FPS\\(Operators)} & \tabincell{c}{Our FPS\\(Scheduler)} \\
        \midrule
        OpenPose (VGG19) & 8\citep{OpenPoseCodes} &19.78 &\textbf{27.32} \\
        OpenPose (MobileNet) & 8.5\citep{tf-pose} &50.89 &\textbf{84.32} \\
        LWOpenPose (ResNet50) & N/A & 38.09 &\textbf{63.52} \\
        LWOpenPose (TinyVGG) & N/A  & 66.62 &\textbf{124.92} \\
        PoseProposal (ResNet18) & 47.6\citep{ChainerPoseProposal} &212.42 &\textbf{349.17}\\
        PifPaf (ResNet50) & 14.8\citep{pifpafcode} & 18.5 &\textbf{44.13}\\
        \bottomrule
    \end{tabular}
\end{table}

\begin{table}
    \caption{Accuracy Evaluation of  Development Platform.}
    \label{table:accu_bench}
    \begin{tabular}[width=\textwidth]{c c c c}
        \toprule
        Configuration & \tabincell{c}{Original\\ Accuracy(map)} & \tabincell{c}{Our\\ Accuracy(map)} \\
        \midrule
        OpenPose (VGG19) & 58.4 & 57.0 \\
        OpenPose (MobileNet) & 28.1 & 44.2 \\
        LWOpenPose (MobileNet) & 42.8 & 46.1 \\
        LWOpenPose (Resnet50) & N/A & 48.2 \\
        LWOpenPose (TinyVGG) & N/A & 47.3 \\
        \bottomrule
    \end{tabular}
\end{table}

\footnotetext{For fair accuracy comparison, the weights are from PifPaf and PoseProposal libraries.}
Our evaluation of \sys is driven by two questions:
(i)~How flexible is its API when developing and customising real-world pose estimation algorithms?
(ii)~How fast is its execution engine in practical deployment environments?

\subsection{API Study}
We compare our API design with existing Python pose estimation libraries: TF-Pose and PyTorch-Pose.
Other libraries such as OpenPose and AlphaPose are dedicated algorithm implementations without customised extensions, we thus exclude them here.

In Table~\ref{tab:api_comparison},
our comparison follows five metrics:
(i)~the number of pre-defined pose estimation algorithms, (ii)~the number of pre-defined datasets,
(iii)~the number of pre-defined backbone deep neural networks (DNNs),
(iv)~the total number of pre-defined
configurations of the pose estimation system, and (v)~the ability
to extend the library to support custom algorithms. 

TF-Pose only supports 1 algorithm, 1 dataset, and 5 DNNs, resulting in up to 5 configurations in its design space.
PyTorch-Pose is more flexible, which supports 2 pose estimation algorithms, 3 datasets, and 13 DNNs, summing up to 26 system configurations. However, PyTorch-Pose only covers single-human scenarios. This is attributed to the insufficient performance of its algorithm execution engine in multi-human scenarios.
Contrastingly, \sys provides 3 algorithms, 2 datasets, and 10 DNNs, thus supporting up to 30 system configurations. Moreover, in all these Python libraries, \sys is the only one that supports the extension of new pose estimation algorithms and provides abstract processing modules that allow users to build their own development pipeline.

\subsection{Performance Evaluation}

Table~\ref{table:perf_bench} compares the existing libraries and \sys.
All benchmarks are evaluated under the same configuration.
The test-bed is of 6 CPU cores and 1 NVIDIA 1070Ti GPU. 
We measure the throughput of the pose estimation systems. 
The benchmark video stream comes from the \textit{Crazy Uptown Funk Flashmob in Sydney} which contains 7458 frames with 640x360 resolution.

We first compare the performance of \sys with the OpenPose framework~\citep{openpose},
which leverages Caffe as its backend and uses C++ for implementing pre-processing and post-processing.
As is shown in Table~\ref{table:perf_bench}, OpenPose is only able to achieve 8 FPS on 1070 Ti, which \sys can reach 27.32 FPS, outperforming the baseline by 3.1x.
On one hand, this improvement is attributed to the careful use of the TensorRT library as the implementation of the inference operator.
On the other hand, the hybrid dataflow operator scheduler makes the execution of \sys even 1.38x faster than the non-scheduled one.
TF-Pose~\citep{tf-pose} leverages TensorFlow as its inference engine and its post-processing is implemented in C++ as well.
When executing MobileNet-based OpenPose, it only achieves 8.5 FPS, which is 10x slower than \sys. 

In addition to OpenPose-based algorithms, \sys also outperforms Chainer's~\citep{ChainerPoseProposal} implementation of Pose Proposal Network by 8 times.
We verified the performance consistency by replacing the backbones and post-processing methods. For example, \sys also beats OpenPose when evaluating a smaller model (\emph{i.e.,} MobileNet).
This proves that the execution engine design is generic so that its benefits should be shared by all custom algorithms. Table~\ref{table:accu_bench} shows the accuracy evaluation result of \sys.

\section{Conclusion}
When operating pose estimation in the wild, developers often
find it challenging to customise pose estimation algorithms for high accuracy,
while achieving real-time pose estimation using commodity CPUs and GPUs. 
This paper introduces \sys, a library for fast and flexible human pose estimation. \sys provides users with expressive APIs for declaring custom pose estimation algorithms. It also provides a high-performance model inference engine
that can efficiently utilise all parallel CPUs and GPUs. This engine enables \sys to achieve 3.1x better performance than existing libraries, while achieving the same accuracy in challenging pose estimation tasks.   

\begin{acks}
This project was supported by National Key R\&D Program of China (2020AAA0103501).
\end{acks}

\bibliographystyle{ACM-Reference-Format}
\bibliography{reference}

\end{document}